# Testing for Feature Relevance: The HARVEST Algorithm

Herbert Weisberg, Victor Pontes & Mathis Thoma

CAUSALYTICS, LLC

**Abstract**

Feature selection with high-dimensional data and a very small proportion of relevant features poses a severe challenge to standard statistical methods. We have developed a new approach (HARVEST) that is straightforward to apply, albeit somewhat computer-intensive. This algorithm can be used to pre-screen a large number of features to identify those that are potentially useful. The basic idea is to evaluate each feature in the context of many random subsets of other features. HARVEST is predicated on the assumption that an irrelevant feature can add no real predictive value, regardless of which other features are included in the subset. Motivated by this idea, we have derived a simple statistical test for feature relevance. Empirical analyses and simulations produced so far indicate that the HARVEST algorithm is highly effective in predictive analytics, both in science and business.

**The Problem**

The main objective of a machine learning algorithm is to generate a highly accurate analytic predictor (model, learning machine) to predict an individual's outcome based on the values of potentially predictive features (variables). There are many different statistical and data mining techniques in common use. These methods are relatively straightforward to apply successfully when a manageable set of relevant features is known in advance. However, in many recent applications, particularly those related to bioinformatics, and to genomics specifically, there may be many more features than there are observations. Moreover, the vast majority of the features probably contribute very little to predictive validity.

In this challenging context, a bewildering plethora of alternative methods has evolved [1,2], but their relative efficacy and ease of use remain unclear There is a need for techniques that can be applied straightforwardly, but are able to overcome the conceptual and practical difficulties entailed in identifying the most relevant features. In this article, we propose a concept of feature relevance and explain how this relevance can be tested. We have termed our method HARVEST, which stands for Highly Accurate and Robust Variable Evaluation Screening and Testing.

The crux of the feature-selection dilemma is the issue of specifying what precisely is meant by feature importance, or *relevance*. Many attempts have been made to craft a working definition in purely statistical terms. We believe this approach to be a dead end. Suppose we can measure a large number of features about some process in a well-defined population of individuals. Each of these features might be thought to provide some predictive information about an outcome of interest $Y$. In a statistical sense, this means the feature is somehow associated with the outcome. However, such an association may be quite subtle and conditional.

In particular, the feature's predictive value may depend on the mathematical form of the model itself and on which other predictive features are included. As there is an unlimited variety of such model specifications, framing a notion of relevance in the abstract seems hopeless. Our proposed methodology is motivated by the notion that in many practical situations, a very large number of potential features are available, but the overwhelming majority are causally unrelated to the outcome of interest. This theoretical "definition" is purely conceptual and cannot be verified in purely mathematical terms. However, it has practical implications.

**Prediction and Causation**

Conceptually, a relevant feature must have a stable statistical relationship with the outcome, i.e. an association that can be expected to persist over time. Clearly, association does not imply causation, an adage drilled into the head of every beginning statistics student. Much less recognized is that a stable association, and hence prediction, does depend on causation in an important sense. The founder of modern statistics, R. A. Fisher, understood this intuitively. In his early foundational work, he wrote that statistical data were generated by an underlying "matrix of causal conditions … of whatever nature they may be" [3].

This *causal matrix* refers to whatever underlying network of interacting causal factors (e.g., physical, biological, psychological, social) determines the value of the outcome. To have any sort of stable statistical relationship with the outcome, a feature must in some way represent, or at least mirror, some aspect of this causal matrix. That is, the feature must measure a factor that is *structurally linked* with the outcome as a result of being part of the relevant causal matrix. Conversely, an *irrelevant* feature lies completely outside the relevant causal matrix. It can be regarded as effectively random with respect to the outcome.

Note that we are not suggesting that a predictive feature must necessarily be a "cause" of the outcome in the usual sense. Rather, we are saying that a stable association, however complex or conditional it may be, must ultimately be grounded in a causal matrix that gives rise to the outcome. Accordingly, we define a *relevant* feature heuristically as one that measures a factor that is structurally linked to the causal pathway generating the outcome. What do we mean by structurally linked?

Perhaps no one in recent years has thought more about causation and prediction in complex systems than Judea Pearl. His preferred mode of exposition is in terms of graphs that represent precisely the kinds of causal matrices I believe Fisher had in mind. He refers to these as *Bayesian networks*. Pearl's pioneering research concerns mainly causation, especially with regard to possibilities for interventions to improve the performance of systems and processes. Whether these Bayesian networks can, in most practical situations, be elucidated in sufficient detail to be of practical use remains arguable. However, this perspective is helpful for understanding the oft-misunderstood relationship between causality and prediction. As Pearl explains, any *stable* associations among a set of features (variables) requires that "if any two variables are dependent, then one is a cause of the other *or* there is a third variable causing both." [4]

In other words, any stable probability distribution describing a set of variables is ultimately determined by a full specification of a causal matrix, or Bayesian network. What this formulation implies is that there are three possible ways in which a feature can be structurally linked with a causal pathway that generates *Y*:

a) The feature itself can be *on* the causal pathway.
b) The feature can be causally influenced *by* a feature on the causal pathway.
c) The feature can be associated with a feature on the causal pathway through a third (confounding) feature.

Furthermore, the causal effects need not be direct, but could be exerted through a chain of intermediate linkages.

Satisfying any of these three criteria renders a particular feature relevant. Whether or not it actually has predictive value depends on which other relevant features are included in the predictive model. As a simple example, assume that $X_1$ has no direct effect on $Y$ but $X_1 \to X_2 \to Y$. Then, if $X_2$ is incorporated in a predictive model, $X_1$ would be superfluous. But if $X_2$ is not available, then $X_1$ would be useful as a predictor. We would consider $X_1$ to be a relevant feature, but redundant in the context of this particular model.

Of course, since most underlying real-world processes are much too complex to be elucidated completely, this definition might appear to be strictly theoretical. How can it possibly assist us in discriminating between relevant and irrelevant features? The answer is that a feature that is *not* relevant in this sense will still have an important statistical property. Such an irrelevant feature can add no real predictive value, regardless of which other features may be included along with it in a predictive model. As we will show, that fact can potentially enable us to distinguish between relevant and irrelevant features.

**Existing Methods for Feature Selection**

Most classical statistical models are predicated on the assumption of a fixed set of available features (independent variables) whose values have been observed on many more individuals than there are variables. In contrast, many practical applications today involve an enormous number of potential features, of which only a few are both relevant and non-redundant. Feature selection can in this case be characterized as the attempt to cull out the few relevant features from among the multitude of available candidates. The methodology proposed in this article provides a straightforward solution to the problem of identifying such relevant features. This methodology may also be useful when the objective is to winnow down a relatively large subset of relevant features to select a smaller combination that is very highly predictive.

Most of the various methods for feature selection attempt to optimize some goodness-of-fit criterion. In theory, this could be accomplished by evaluating this criterion for every possible subset of features. However, a practical combinatorial problem arises because of the astronomical number of possible subsets of the predictive features. This "curse of dimensionality" renders the brute-force approach computationally infeasible. Instead, it is necessary to *search* for a solution efficiently, while testing only a very small fraction of all the various possibilities. Furthermore, even an exhaustive search would not necessarily identify the relevant features correctly, because the accuracy can only be estimated based on the available data. Unless the sample of observations is very large, these estimates may be subject to substantial random error.

For example, genomic studies today are usually based on at most a few hundred individuals. Typically, the goal is to identify a gene *signature* of between about 10 and 100 genes whose expression levels are highly predictive. Overfitting may result from the fact that so many combinations of features are being tested to train (fit) models on this rather modest sample of observations. Consequently, some subsets of features are bound to fit the training data well (i.e., produce good predictions) largely by chance. A model that incorporates these apparently relevant features will often perform poorly in independent validation.

There is an inherent tension between model training and feature selection. For a fixed set of features, it is generally optimal to fit the data as closely as possible. Therefore, in statistical terms, we generally attempt to minimize residuals or maximize the likelihood. However, with many features to choose from, this "best-fitting" strategy breaks down. Increasing the number of features can always increase the observed fit, but overfitting will yield features that are not truly relevant (false positives). There is no perfect solution to this problem, although various methods of cross-validation and testing in an external dataset can be employed to mitigate it [5].

There are three general approaches to the problem of searching for relevant features [6]. These vary according to how closely the process of feature selection is tied to the process of *training* the algorithm:

> Feature selection methods are usually classified into three categories: *filter methods* select subsets of variables as a pre-processing step, independent of the chosen predictor; *wrapper methods* utilize the learning machine of interest as a black box to score subsets of variables according to their predictive power; finally, *embedded methods* perform variable selection in the process of training and are usually specific to given learning machines.

In short, filter algorithms generally do not utilize a statistical model, wrapper algorithms utilize a model primarily as a vehicle for finding relevant features, and embedded algorithms generate relevant features in conjunction with training the model.

A simple example of a filter method would be to calculate a univariate correlation (or other appropriate measure of association) between each feature and the outcome. Then, all those features that exceed a pre-specified cutoff might be selected. Filter methods are simple and fast, and can often serve as a preliminary screening tool. One major drawback is that a high apparent predictive accuracy is often attributable to chance, especially in small samples. A second is that univariate predictive accuracy often fails to translate into multivariate relevance, because a relevant feature may only add significant value when certain other features are also included.

A wrapper method typically involves an algorithm for searching through the enormous space of possible combinations of features. The objective is to find a subset of features that produces the highest estimated predictive accuracy. The accuracy of each subset is evaluated using a specified type of model, such as ordinary least-squares (OLS), logistic regression, or support vector machines (SVM). Once a final set of features has been determined, it may be used in several different ways to derive a final learning machine. As a simple example, if a linear OLS regression model has been used for feature selection, interaction terms or transformed versions of these features may be added in a final training process, ideally performed on a fresh set of data.

Simple examples of wrapper methods are various "greedy" forward and backward selection techniques that have been employed by statisticians for decades. Genetic algorithms, which have more recently become popular, involve a similar approach, but are far more computationally intensive [7]. Sequential search strategies entail iterative steps in which features are added or subtracted from the model. The aim is to continually improve the accuracy of the resulting predictor. The criteria for adding or subtracting features can depend on criteria that are set by the data analyst. However, these criteria can be arbitrary and do not guarantee that the final subset of features will be in any sense close to optimal. Moreover, sequential search algorithms are susceptible to overfitting and possible convergence toward a local maximum in the space of possibilities.

With embedded methods, the processes of feature selection and model training are completely merged. The training process that is used to construct the learning machine generates a presumably relevant subset of features as a byproduct. Examples of such methods are neural networks, SVM, Lasso, elastic net, *k*-nearest neighbor (kNN), and recursive partitioning (decision tree) methods such as CART and Random Forests. These methods were developed in part to overcome the limitations of wrapper methods, especially their penchant for overfitting to the training data. There are two general strategies employed in embedded methods to reduce overfitting.

One approach involves the modification of regression models by imposing a "penalty" on coefficients to "shrink" them back towards zero. Techniques like Lasso and elastic net that apply this approach are somewhat conservative, in the sense that extreme OLS coefficients are given substantially reduced weight [8,9]. This may be an advantage in the context of high-dimensional data in which there are sets of features that are highly inter-correlated, which can lead to inflated and unstable coefficients. Furthermore, the shrinkage can be engineered to reduce many of the smaller coefficients to zero, thereby eliminating them from the model. This is useful in genomic applications, since it is probable that very few genes are relevant.

A second common strategy involves the so-called *ensemble* approach. Rather than building a single learning machine, an algorithm is trained on several randomly selected subsets of the training data. Then, an aggregate predictor combines the results for this whole group of subsets. The basic idea is to achieve greater model stability and thus reduce overfitting. A popular example of the ensemble strategy is the Random Forest method, in which a number of decision-trees are derived [10]. Then, the outcome value (or class) for any new individual is based on an average of all the predictions from individual trees. Such an approach entails greater complexity and computation than a straightforward wrapper method. Moreover, it may be much more difficult to interpret the predictive model or to assess the relevance of individual features.

**Rationale for an Alternative Approach**

While the theories underlying many wrapper and embedded techniques may be plausible, it is not clear to what extent their application has been successful in practice. There has been a proliferation of different specific methods, with no consensus regarding which ones function best in general or in particular situations. The complexity of many methods, along with the "hand-crafting" required to implement them make it difficult to compare their performance. In particular, these techniques often require the analyst to choose values for one or more "tuning parameters." These choices are somewhat arbitrary and empirically based. Thus, it is usually difficult to disentangle the performance of the technique from the proficiency of the analyst.

HARVEST can be considered a type of wrapper, in that it utilizes a predictor to evaluate the potential relevance of various features. However, its objective is that of a filter, since it seeks merely to screen out irrelevant features. Those demonstrably relevant features that remain after this winnowing process can then be subjected to a final round of feature-*selection*. This ultimate analysis would be aimed at discovering an optimal combination of features from among the greatly reduced set of all relevant features. By clearly separating the *testing* of features for relevance from the *search* for an optimal subset of features, the method avoids the traps of overfitting and convergence to a local maximum.

To understand the rationale for HARVEST, let us assume that we have selected a general class of predictive models, such as linear regression, regularized regression or recursive partition. For each such approach, we will have an associated measure of a given model's true accuracy $A$, such as $R^2$ for a continuous outcome or the AUC for a binary outcome. This true accuracy is assumed to be the highest accuracy that could be attained for any predictive model of a given class (e.g., general linear models with first-order interactions). $A$ is the accuracy that would be achieved in an effectively infinite sample of observations. Of course, for any actual situation, we will have only an estimate of $A$ based on a finite sample of observations. Let $\hat{A}$ represent the estimated accuracy based on the observed data. Furthermore, we assume for convenience that $A$ is scaled so that a value of zero corresponds to complete statistical independence between the outcome and the model's predictions.

Assume that there are $p$ available features, all of which are irrelevant in the sense defined above. That is, none of these features are in fact structurally associated with a causal pathway that leads to the outcome we wish to predict. Under this null hypothesis, the value of $A$ must be zero for any model that includes only a subset of the features. Now suppose that some of the features are in fact relevant. Then, a model based on a feature-subset that happens to include one or more of these relevant features would have $A > 0$.

Now suppose we have a subset of size $k$ for which $A > 0$ and we replace an irrelevant feature with a relevant one. Then, such a replacement *might* increase the value of $A$ but can never decrease this value. To understand why, it is helpful to envision again an underlying causal network. The values for most of the relevant features, and their structural linkages, may be hidden. We can think of the $k$ features included in a predictive model as providing a partial representation of the full network. The corresponding value of $A$ indicates the amount of predictive information provided by this partial view of the network.

By observing $k$ features we "reveal" probabilistic information about their structural linkages with the outcome $Y$. However, if a particular feature is irrelevant, then the value of $A$ essentially ignores this feature and would not change if it were removed. Now, suppose we remove such a non-informative feature but add back into our predictive model a relevant feature to replace it. Then, this new feature would uncover any previously hidden stable associations among itself, the other $k - 1$ features, and $Y$.

Recall that we assume $A$ is the optimal accuracy for any model of the given type. Then, there are two possibilities: that the added relevant feature allows the optimal value of $A$ to be improved, or that the added feature can be ignored, since it provides no new useful information. In the latter case, the value of $A$ will depend only on the $k - 1$ features, just as in the original set of $k$ features. For example, the new feature might be linked to a causal pathway only through its effect on a downstream feature already included among the $k - 1$ features. Thus, replacing an irrelevant feature with a relevant feature can never make things worse. Conversely, replacing a relevant feature with an irrelevant one can never make things better.

Now, let us consider all $\binom{p}{k}$ possible subsets of size $k$. Of these, there are $\binom{p-1}{k-1}$ that include a particular relevant feature and $\binom{p-1}{k}$ that do not. All of the $k$-subsets that do not include this feature can be obtained by replacing this feature with one of the other $p - k$ features. In general, this replacement can either an increase or decrease the value of $A$. However, in the situation of most practical interest, the proportion of relevant features will be very small, so, in the great majority of cases, the relevant feature will be replaced by an irrelevant feature.

Suppose that we randomly select a subset that contains a particular relevant feature $r$ and a subset that does not. Let $A_r$ represent the accuracy of the former and $A_{-r}$ the accuracy of the latter. Then, we would have:

$$P(A_r \geq A_{-r}) > .5$$

This statement has a purely combinatorial basis. Therefore, it depends solely on our assumption about the small proportion of relevant features. However, it pertains to the true accuracy values. In practice, we would have only estimates $\hat{A}$ based on available data.

If the sample of observations is very large, the distribution of the estimate accuracy values may be such that:

$$P(\hat{A}_r \geq \hat{A}_{-r}) > 0.5$$

Of course, the value of $P(\hat{A}_r \geq \hat{A}_{-r})$ will depend not only on the true values of $A_r$ and $A_{-r}$ but also on the volume of observed data. With modest sample sizes, this probability may be only slightly greater than 0.5, or even less than 0.5. As the number of observations increases, this probability will increase, and eventually approach its value if the true accuracies were known.

The foregoing logic suggests an obvious way to test whether any particular feature $i$ is in fact relevant or not. Suppose that we first randomly draw a large sample of subsets, each of size $k$, all of which contain feature $i$. Then, we draw a "control" sample of size-$k$ subsets that do not contain this feature. For each subset in both samples we could calculate the accuracy $\hat{A}$. We could then form all pairs of subsets, with one containing feature $i$ and one not containing it. Under the null hypothesis that feature $i$ is truly irrelevant, the probability would be $\leq 0.5$ that the $\hat{A}$ of the subset containing feature $i$ would be greater. (It would be exactly 0.5 if all $p$ features were irrelevant.) Therefore, the proportion of such pairs in which the subset containing feature $i$ has greater accuracy is a test statistic for this null hypothesis.

Statisticians will recognize that what has just been described is in fact an application of the well-known Mann-Whitney test. This test is mathematically equivalent to a simpler and more common version known as the Wilcoxon Rank Sum test. The Wilcoxon version in this context requires a simple ranking of all the $\hat{A}$ values in the two samples of subsets combined [11]. Then, it is only necessary to calculate the average *rank* of the subsets that contain the feature being tested.

Under the null hypothesis, the exact probability distribution of the average rank is known. This distribution can be used to test whether the average rank for the subsets containing a particular feature is truly higher than would be expected under the null hypothesis. Using the Wilcoxon Rank Sum test in this way to check the statistical significance for each feature is at the heart of our proposed algorithm for feature selection. However, before describing the HARVEST algorithm, a *caveat* is in order.

From our discussion above, it might seem that we could achieve unlimited increases in statistical power, while maintaining Type I error control, simply by increasing the size of the random samples of size-$k$ feature subsets. However, the calculation of $\hat{A}$ for each size-$k$ subset is conditioned on the same finite sample of underlying observations. Therefore, any idiosyncrasies of this particular sample will affect the relative rankings of various subsets. Indeed, the values of $\hat{A}$ may not be sufficently close to those of $A$ for $P(\hat{A}_r \geq \hat{A}_{-r})$ to hold.

**The HARVEST Algorithm**

Suppose that we select $\lambda$ random subsets that contain a feature being tested, and a control sample of $n$ random subsets without this feature. To perform a separate Wilcoxon test for each of the $p$ possible features would require evaluating $2\lambda p$ subsets. For example, with 10,000 features and 50 subsets with and without feature $i$, it would be necessary to train 1,000,000 base models. Fortunately, there is a much more efficient way to perform the necessary calculations. Rather than a separate test for each feature, we simply draw one large sample of subsets, and allow each feature to be sometimes the "test subject" and sometimes the "control" for some other features.

Assume we have available a training sample of $N$ observations randomly drawn from the population of interest. For each observation, the values of $p \gg N$ features are known. These features, can have any form (e.g., binary, categorical, continuous). Let $A$ be an appropriate criterion that measures accuracy (e.g., $R^2$, $AUC$, $AIC$) for the predictor (model) estimated using the training sample. The general form of the model (e.g., OLS, logistic, Lasso etc.) must be specified.

Our objective is to identify all relevant features, while excluding most irrelevant features. For example, suppose there are approximately 50 relevant features among a total of 1,000. We may be willing to tolerate finding around 10 false positives in order to be highly likely to confident of identifying at least 45 of the relevant features. So, we could accept a Type-1 error of 0.01. The HARVEST algorithm starts by specifying this p-value, and then tests every feature as follows:

Step 1: Establish an acceptable false-positive error rate (p-value).
Step 2: Generate $n$ randomly-selected subsets of size $k$ out of all $\binom{p}{k}$ possible subsets.
Step 3: For each random subset, train the corresponding predictive model.
Step 4: For each random subset, calculate an estimate of accuracy $\hat{A}$ in the training sample.

Step 5: Rank all *n* subsets based on their values of $\hat{A}$ from 1 (highest) to *n (*lowest).

Step 6: For each feature *i*, identify the $n_i$ subsets that contain feature *i*.

Step 7: For each feature *i*, calculate the *average rank* of the $n_i$ subsets that contain the feature.

Step 8: For each feature *i*, apply the Wilcoxon Rank Sum test to calculate the p-value of this average rank.

Step 9: Eliminate from consideration those features that fail to achieve the significance level specified in Step 1.

Once the HARVEST algorithm has "harvested" the significant features, they can be used in a variety of ways. For example, they can be used to train a learning machine, preferably on an external validation sample. Other well-established selection methods, such as stepwise regression procedures, can possibly be applied. Making data transformations or adding interactions can be employed to refine the model. Or, these significant features can be used to train a different type of predictive model, such as a decision-tree. In any case, the validity of this final model should be tested in a hold-out sample or a completely independent dataset.

Another strategy is to use the resulting significant subset as input to a second, and possibly additional rounds of HARVEST. In the subsequent rounds, there may be a majority of relevant features, although some may be "weaker" predictors than others. The objective would be to distinguish the "stronger" features from the weaker ones. In terms of the causal network conceptualization, we might think that a strong feature would tend to lie close to *Y* along a causal pathway. Such a feature's causal effect on *Y* might not only be strong, but also less likely to be obviated by that of another downstream feature already in the model.

**Implementing HARVEST**

Although the HARVEST algorithm can be run in an automated manner, five "settings" must be specified by the user:

- The general form of the learning machine
- The specification of a desired p-value
- The accuracy criterion: *A*
- The total number of random subsets: *n*
- The number of features included in each subset: *k*

The choice of machine-learning approach and accuracy criterion are a matter of individual preference. In our research so far, we have used OLS regression for numerical outcomes and logistic regression for binary outcomes. As corresponding accuracy criteria we have used $R^2$ and $AUC$ respectively. So far, we have experimented with values of 10, 15 and 20 for $k$, and p-values of 0.05 and 0.01, with or without Bonferroni adjustment or the less conservative false discovery rate approach [12]. When feasible, a sufficient total number of random subsets have been selected to allow each feature to be included in approximately 100 subsets.

The statistical power, as well as the computational burden, will depend on these settings. In general, it is impossible to calculate the statistical power, because a precisely-specified alternative hypothesis cannot be defined. However, in general it will be true that increasing the values of $n$ and $k$ will increase the statistical power, since larger values of $n$ and $k$ will result in a larger number of subsets $n_i$ that include a given feature.

For any particular feature, the probability is $k/p$ of being contained in any random subset. Therefore, out of $n$ subsets (sampled with replacement), the number $(n_i)$ of them that contain this feature has a binomial distribution with:

$$E(n_i) = \frac{nk}{p}$$

However, since $k$ is usually very much smaller than $p$, the distribution of $n_i$ is approximately Poisson with both a mean and variance $nk/p$. This formula can be used to determine a value of $n$ that will be considered adequate. For example, suppose we have 1,000 possible features and 20 of these are included in each size-$k$ subset. Then to obtain approximately 100 subsets (with a standard deviation of 10) for each feature would require a total of 5,000 subsets.

In Step 8 of HARVEST, we mentioned that the Wilcoxon Rank Sum test can be used to calculate the p-value for each feature $i$. Adapting a well-known approximation leads to a simple expression for the null distribution of the average rank of the subsets that contain feature $i$. *This average rank is approximately normal with mean $\mu_i$ and standard deviation $\sigma_i$:*

$$\mu_i = \frac{n+1}{2}$$

$$\sigma_i = \sqrt{\frac{(n-n_i)(n+1)}{12n_i}}$$

From these formulas, we can easily compute the p-value for any observed value of the average rank. Note that when utilizing HARVEST this test should be "one-sided" because it is only possible for a feature to be *more* predictive than a randomly selected feature, not less.

**A Genomic Example**

We recently completed a test of HARVEST as a method for identifying a genomic signature. This test was conducted in conjunction with GenomeDx, which developed and markets a diagnostic procedure called Decipher. This product is one of several that have recently been developed to improve the accuracy of prognosis for prostate cancer patients [13]. In particular, Decipher can assist in deciding whether additional treatment is required for prostate cancer patients who have undergone a radical prostatectomy. The primary endpoint predicted by Decipher is occurrence of metastasis within five years after prostate removal. Improving the physician's ability to forecast this endpoint is important, because decisions about possible additional treatment can hinge on the probability of eventual metastasis.

Decipher is based on a *genomic classifier* (GC), a predictive model derived on the basis of data provided by the Mayo Clinic [14]. The data pertained to a sample of 545 patients. Of these, 359 were utilized for classifier training, and 186 were held out for internal validation. Subsequently, the GC was validated further on several external datasets obtained from major clinical centers. The final GC is based on the expression levels for 22 features, including some non-coding RNAs as well as protein- coding genes. These 22 were chosen from a total pool of over 1.4 million features using a complex combination of sophisticated machine-learning approaches. An initial univariate filtering step was followed by Lasso regression with bootstrapping to obtain a set of 43 candidate features. The final model was based on a Random Forest analysis. This analysis included a tuning process that involved a cross-validation step to winnow the feature set down to the 22 ultimately selected.

The accuracy of the Decipher GC was assessed in several ways, with AUC being the primary metric of interest. The value of AUC in the internal validation set was 0.75, higher than any previous accuracy reported for predicting future metastasis in the post-prostatectomy population. The median AUC value achieved in five subsequent external validations was also 0.75. Thus, Decipher is a well-validated predictor that represents the current gold standard for predictive accuracy in this population of patients. The purpose of our test was to determine whether a simpler approach based on HARVEST could obtain comparable or better results.

The test of our methodology utilized gene expression data provided by GenomeDx for approximately 6,000 genes. The dataset for all 545 of the patients underlying its initial model derivation were provided. We decided to apply HARVEST with logistic regression as the form of the learning machine and $AUC$ as our criterion of accuracy. We used the GenomeDx training sample of 359 observations to train our model. We applied HARVEST on a random 80% sample of the 359, and held out the remaining 20% for validation. After three successive rounds of HARVEST, we obtained a final set of 10 highly relevant genes.

We included this set in a simple linear regression model and obtained an $AUC$ of 0.83 in the training sample, and, when applied to the internal validation data, an $AUC$ of 0.68 resulted. We then used the 10 genes produced by HARVEST, and fit the final logistic model using the 278 observations (20% holdout and internal validation) that had not been used in the feature selection. The intention was to mitigate the problem of overfitting, while still allowing us to validate on the five external patient samples.

These final 10 features (genes) selected by HARVEST were than submitted to GenomeDx. The expression data for these genes for the patients in the five external validation studies were then returned to us. Using our derived model, we then scored all the validation patients and produced an estimated metastasis probability for each patient. GenomeDx then calculated the $AUC$ and other measures of accuracy for these estimates. The results were extremely encouraging as our model produced a median $AUC$ of 0.79 for the five external validation samples, compared with 0.75 for Decipher.

**Simulation**

We recently completed several simulation analyses to test the performance of HARVEST. So far, these simulations have assumed that the underlying true models that generated the training data were GLM. The results have been very successful. We describe here a simple model that has previously been analyzed using several well-known regularized regression techniques. We compare the results attained by HARVEST with those achieved by these other methods. The data-generating model and results are presented in Wang et al [15]. These authors compared their own proposed *random lasso* algorithm to four other prominent methods: *elastic net* [9], *adaptive lasso* [16], *relaxed lasso* [17], and *VISA* [18].

The model that we have chosen is Example 4 in Wang et al., which is a simple linear regression with a zero intercept and 40 independent variables. Two simulations were performed, with sample sizes of 50 and 100 respectively. Each of the independent variables in the model was Normal with mean 0 and variance 1. The first six of these independent variables had coefficients with the values 3,3,-2,3,3,-2 respectively, and the remaining 34 had zero coefficients. The pairwise correlations among the first three and among the second three variables were 0.9. The 34 irrelevant variables were independent of each other. The first set of three variables, the second set of three variables, and the remaining 34 variables were independent of each other. The error variance was 36.

HARVEST was applied with the following settings:

- General model form: simple linear regression
- Desired p-value: 0.05
- The accuracy criterion: $R^2$
- The total number of random subsets: 4,000
- The number of features included in each subset: 15

In [15], each of the two simulations (sample sizes 50 and 100) was replicated 100 times. The authors then found for each of the 40 variables, how many times that variable was selected by each of the algorithms. For each of the 6 relevant variables, the number of times selected can be interpreted as an estimate of the method's sensitivity. For each of the 34 irrelevant variables, the number of times not selected provides an estimate of the specificity. Table 2 in [15] presents (in effect) the minimum, median and maximum values of these 6 sensitivity values and 34 specificity estimates.

Our Table 1 displays the results in [15] for the six techniques they compared, along with the corresponding results for HARVEST based on our simulation. HARVEST outperforms all six competitors, especially for its intended purpose of ensuring that virtually all relevant features are retained. Even with only 50 observed individuals, HARVEST achieved very high sensitivity (minimum of 93%) for all 6 relevant variables. Meanwhile, its specificity was comparable to that of the other standard techniques. In other words, this excellent sensitivity was not obtained by sacrificing specificity.

**Discussion**

Feature selection with high-dimensional data and a very small proportion of relevant features poses a serious challenge to standard statistical methods. Existing methods are complex, difficult to apply, and often unsuccessful. We have developed an alternative approach that has a compelling theoretical rationale and is relatively straightforward to apply. The basic idea is to evaluate each feature in the context of many random subsets of other features. HARVEST assumes that in many practical situations a relevant feature is likely to add real predictive value to at least some of these subsets. In contrast, the irrelevant features will add no real value and can only *appear* to be associated with the outcome by chance. If this assumption holds, then subsets will tend to achieve a higher level of predictive accuracy to the extent that they contain more and/or stronger relevant features. Capitalizing on this intuitive idea allows us to perform a statistical test of significance via the ranking of a single large sample of equal-sized subsets.

While it represents a unique new approach, HARVEST shares some aspects with other recently proposed methods. In particular, the use of random subsets of features has been explored quit extensively, primarily in the context of recursive partition models. Ensemble approaches, most notably random forests, include random subsets of features within the tree-growing algorithm. Other "random subspace" methods typically employ the random subsets as part of a sequential search algorithm [19].

Random lasso, entails generating multiple models based on random samples of features. However, the aim is to identify important features by essentially averaging the estimated coefficients across the different models. We are aware of only two proposed methods that are similar to HARVEST in the sense of explicitly evaluating the relevance of individual features according to their degree of participation in highly-performing random subsets of features [20, 21]. Both of these techniques involve sequential search and have been developed in the context of the kNN algorithm.

Finally, there are many questions regarding how and in what circumstances HARVEST can best be utilized. In particular, the optimal settings for the total number of model's $n$ and the number of features per model $k$ in various types of situations are not obvious. Our research has assumed only OLS regression or logistic regression for the type of learning machine. In principle, there is no reason not to try more sophisticated and computationally complex algorithms. Similarly, different accuracy measures could be tried.

**Conclusion**

To conclude, most of the usual approaches to feature selection with high-dimensional data try to optimize a criterion of model performance in the training data. As a result, a choice must be made between Type I and Type II errors. To avoid false discovery, a price must be paid in over-conservatism. We have proposed a possible way to resolve this dilemma by generating a large sample of random subsets of features. Then, the performance of a feature can be assessed across the various subsets in which it appears. Our experience indicates that HARVEST can be useful in predictive analytics, both in science and business.

**Table 1**

**Comparison of HARVEST with Six Other Methods (n = 50)**

|  | Sensitivity | | | Specificity | | |
|---|---|---|---|---|---|---|
| **Method** | Minimum | Median | Maximum | Minimum | Median | Maximum |
| Lasso | 11 | 70 | 77 | 75 | 83 | 88 |
| Adaptive Lasso | 16 | 49 | 59 | 86 | 92 | 96 |
| Elastic Net | 63 | 92 | 96 | 77 | 83 | 91 |
| Relaxed Lasso | 4 | 63 | 70 | 91 | 96 | 100 |
| VISA | 4 | 62 | 73 | 92 | 97 | 99 |
| Random Lasso | 84 | 96 | 97 | 70 | 79 | 89 |
| **HARVEST** | **93** | **95** | **98** | **84** | **91** | **96** |

**Table 2**

**Comparison of HARVEST with Six Other Methods (n = 100)**

|  | Sensitivity | | | Specificity | | |
|---|---|---|---|---|---|---|
| **Method** | Minimum | Median | Maximum | Minimum | Median | Maximum |
| Lasso | 8 | 84 | 88 | 69 | 78 | 88 |
| Adaptive Lasso | 17 | 62 | 72 | 86 | 90 | 96 |
| Elastic Net | 70 | 98 | 99 | 79 | 86 | 93 |
| Relaxed Lasso | 3 | 75 | 84 | 92 | 97 | 99 |
| VISA | 3 | 76 | 85 | 91 | 96 | 99 |
| Random Lasso | 89 | 99 | 99 | 79 | 86 | 92 |
| **HARVEST** | **94** | **99** | **100** | **90** | **96** | **99** |